\documentclass[11pt]{article}

\usepackage[margin=1in]{geometry}
\usepackage[T1]{fontenc}
\usepackage{times}
\usepackage{microtype}
\usepackage{amsmath,amssymb}
\usepackage{booktabs}
\usepackage{array}
\usepackage{multirow}
\usepackage{graphicx}
\usepackage{xcolor}
\PassOptionsToPackage{hyphens}{url}
\usepackage{hyperref}
\usepackage{natbib}
\usepackage{tikz}
\usetikzlibrary{arrows.meta,positioning,fit,calc}
\usepackage{caption}
\usepackage{enumitem}
\usepackage[utf8]{inputenc}
\usepackage{listings}
\newcommand{\local}{\textsc{local64}}
\newcommand{\bpe}{\textsc{BPE-8k}}
\newcommand{\bpb}{\mathrm{BPB}}

\title{Byte Patches versus {BPE} at Matched Data:\\
A Matched-Twin Comparison at {350M} Parameters}
\author{Asadullah Maliyawala \and Andrej Lisal\\independent researchers}
\date{}

\begin{document}
\maketitle

\begin{abstract}
We compare a MEGABYTE-style byte-patch Transformer with a 64-byte sliding local
decoder (\local) to a matched byte-level 8k BPE decoder (\bpe) under identical
raw bytes, document order, and approximately matched bytes per optimizer step:
350M parameters, $10^{10}$ bytes, real\_v3.
Under matched-span document sliding (identical scored byte ranges;
\bpe{} $\min(512~\mathrm{tokens},\,2048\,\mathrm{B})$),
\bpe{} leads overall/text/code/math with relative gaps
$-5.2\%$, $-4.9\%$, $-9.3\%$, $-6.1\%$
(overall, text, code, math; paired bootstrap).
Downstream, CharBench equal fine-tuning favors \local{}
($53.7\%$ vs~$40.0\%$), fact200 favors \bpe{} ($16.5\%$ vs~$7.0\%$),
and PIQA (raw and length-normalized) and ARC-Easy length-normalized favor \bpe{} after Holm
correction at $\alpha=0.05$; remaining tasks are ties.
Evaluation pitfalls that changed conclusions include context-length
extrapolation past the BPE pack, bits/token reported as BPB, and inconsistent
byte denominators.
Unlike controlled tokenizer studies that vary the subword tokenizer under a
fixed token budget, or FLOP-controlled byte-model scaling studies, we hold the
raw bytes, their order and the optimizer recipe fixed and ask whether a
fixed-patch byte model can replace the tokenizer at all.
Limits: one size, one seed, one 8k vocabulary fixed before the runs, and a known difference in embedding initialization between the twins (Section~\ref{sec:limits}).
\end{abstract}

\section{Introduction}
\label{sec:intro}

Byte-pair encoding (BPE) and related subword tokenizers~\citep{sennrich2016}
are the default interface between raw text and modern language models.
They compress frequent strings into atomic vocabulary items, reducing sequence
length, but fragment spelling and morphology and make code whitespace brittle.
A tokenizer is also a fixed, separately trained component: it is chosen before
the language model, it is specific to the data it was fitted on, and the model
cannot revise it.

An alternative is to model bytes directly and recover efficiency with
hierarchical structure.
Operating on bytes removes the tokenizer and gives the model access to every
character, at the price of much longer sequences.
Hierarchical designs address that price by letting a large, expensive model
operate on short sequences of coarse units (here, groups of four bytes) while a
small model fills in the individual bytes.
We study a MEGABYTE-style~\citep{megabyte} variant: a patch-4 encoder, a global
Transformer, and a two-layer causal byte decoder whose 64-byte attention window
crosses patch boundaries (\local).
The baseline is a decoder-only Transformer over an 8k byte-level BPE vocabulary
(\bpe).

Claims in this area are easy to get wrong, because byte models and token models
are scored in different units, see different amounts of context, and are
trained with different sequence lengths.
Recent work makes the same point from the tokenizer side.
Controlled studies show that the choice of subword tokenizer changes downstream
accuracy and cost~\citep{ali2024tokenizer,schmidt2024compression,meister2026tokeval},
that the best vocabulary size depends on the compute budget~\citep{tao2024vocab},
and that bits-per-byte (BPB), the usual cross-tokenizer metric, is itself not
tokenizer-agnostic and can disagree with task accuracy~\citep{schaeffer2026myths}.
These studies compare tokenizers with one another, mostly under a fixed token
budget.
At large scale, byte models with learned patches have been compared to
tokenized models under FLOP control~\citep{blt}.
What is missing is a small, tightly matched answer to a simpler question:
\emph{if the raw data, its order and the training recipe are held fixed, does a
fixed-patch byte model keep up with a BPE twin, and which evaluation choices
change the answer?}

We therefore build the comparison around \emph{matched twins}: two models of
the same width and depth, trained on the same bytes in the same order with the
same optimizer recipe, and then evaluated so that both score the same byte
ranges with the same left context.
Holding the budget in \emph{bytes} rather than tokens matters here: under a
token budget, the tokenizer with better compression sees more raw text, which
confounds the architecture comparison.

\paragraph{What is different about this study.}
Three design choices separate this work from the studies above
(Table~\ref{tab:related}).
(i)~\emph{The unit of matching is the raw byte stream.} Both models are trained
on the same $10^{10}$ bytes, in the same document order, with a learning-rate
schedule keyed to bytes seen, so neither model gets more text or a different
curriculum; the remaining compute difference is about $2\%$ and favors the byte
model.
(ii)~\emph{The comparison removes the tokenizer instead of swapping it.}
Controlled tokenizer studies change \emph{which} subword tokenizer a fixed model
uses; we ask whether a fixed-patch byte model can do without one, at matched
depth and width.
(iii)~\emph{Evaluation is treated as part of the result.} We score both models
on identical byte spans with identical left context, and we report how much the
conclusion moves when that control is dropped.
The size of that effect is the main warning of this paper: the \emph{same}
\bpe{} checkpoint scores $0.600$ bits per byte on code under matched spans but
$0.843$ when it is asked to score 2048-byte windows past its trained context,
which turns a clear BPE lead on code into an apparent byte-model win
(Section~\ref{sec:pitfalls}, Table~\ref{tab:uncapped350}).
This is a concrete, measured instance of the concern raised
by~\citet{schaeffer2026myths}: cross-tokenizer BPB comparisons are only as
meaningful as the evaluation protocol behind them.

\paragraph{Question.}
Under matched data---same raw bytes, same document order, same optimizer recipe,
and approximately the same bytes per optimizer step---does \local{} match or beat
\bpe{} on byte-normalized likelihood and downstream tasks at 350M / $10^{10}$ bytes?

\paragraph{Answer in brief.}
No on likelihood: when both models score identical byte ranges, the BPE twin has
lower bits-per-byte in all three domains we measure (text, code, math), by
$4.9$--$9.3\%$ (Section~\ref{sec:results}).
On downstream tasks the picture is mixed: the byte model wins a character
manipulation benchmark after equal fine-tuning, the BPE model wins a factual
recall set and two multiple-choice sets, and the remaining tasks are ties
(Table~\ref{tab:down}).
A second finding is methodological: the sign of the code result changes with
the evaluation protocol, and several evaluation errors that we found and fixed
would each have changed a headline number
(Section~\ref{sec:pitfalls}).

\paragraph{Contributions.}
\begin{itemize}[leftmargin=*,itemsep=2pt]
\item \textbf{A byte-matched twin design.} Matched twins at 350M on shuffled
real\_v3 ($10^{10}$ bytes): same byte stream order, AdamW recipe, and retuned
bytes/step, so the comparison is data-matched (training compute is within
$\sim$2\%; Table~\ref{tab:configs}).
Table~\ref{tab:related} contrasts this design with the closest prior studies.
\item \textbf{A context-controlled likelihood result.} Under matched spans
(Table~\ref{tab:bpb}), \bpe{} leads overall/text/code/math, alongside a
downstream comparison that does not rely on BPB (Table~\ref{tab:down}).
\item \textbf{Measured evaluation pitfalls.} An evaluation checklist covering
BPB denominators, first-position asymmetry, LAMBADA targets, and context-length
extrapolation (Section~\ref{sec:pitfalls}), with the measured effect on our
numbers where we have one; it complements analyses of why BPB is not
tokenizer-neutral~\citep{schaeffer2026myths} with concrete errors that flipped
results in a byte-versus-BPE setting.
\item Released code, evaluation suites with hash locks, and checkpoints
(Appendix~\ref{app:repro}).\footnote{Code, evaluation suites, checkpoints and validation data: \href{https://github.com/AsadullahMaliyawala/bytelm-local64}{\texttt{github.com/AsadullahMaliyawala/bytelm-local64}} (full list in Appendix~\ref{app:repro}).}
\end{itemize}

\paragraph{Roadmap.}
Section~\ref{sec:related} places the work among related models.
Section~\ref{sec:model} specifies both models and Section~\ref{sec:setup} the
training setup.
Section~\ref{sec:protocol} defines the evaluation protocols, and
Sections~\ref{sec:results}--\ref{sec:analysis} give results and discussion.
Section~\ref{sec:pitfalls} lists the evaluation pitfalls, and
Section~\ref{sec:limits} states the limitations.

\section{Related work}
\label{sec:related}

\paragraph{Subword tokenization.}
Subword units learned by byte-pair encoding~\citep{sennrich2016} balance
vocabulary size against sequence length and are the standard input
representation.
Our baseline uses a byte-level BPE vocabulary of $8000$ items, so that every
input string is representable.

\paragraph{ByT5.}
\citet{byt5} train encoder--decoder models on raw UTF-8 bytes.

\paragraph{MEGABYTE.}
\citet{megabyte} introduce multiscale Transformers that predict byte patches
with a global model and expand locally.
Their local model decodes one patch at a time, conditioned on the global state,
so local attention stays inside the patch.
Ours uses a two-layer causal decoder whose 64-byte window crosses the fixed
4-byte patch boundary, and the comparison is against one matched BPE twin rather
than a scaling law over patch size (Appendix~\ref{app:arch}).
We do not reimplement MEGABYTE; ``MEGABYTE-style'' refers to the global/local
division of labor only.

\paragraph{Byte Latent Transformer (BLT).}
\citet{blt} dynamically form latent patches from bytes with an entropy model and
report a FLOP-controlled scaling study up to 8B parameters and 4T training bytes,
finding that byte patches can match tokenization-based models at scale.
Their comparison controls compute across a scaling sweep; ours controls the
data itself (identical bytes in identical order) for one pair of models.
We keep a fixed patch of four bytes and train a BPE twin on the same documents
in the same order at 350M, so the two results are complementary rather than
competing: BLT asks how the trade-off scales, we ask what a single tightly
matched pair shows and which evaluation choices change it.

\paragraph{Other byte / character models.}
Hourglass~\citep{hourglass}, SpaceByte~\citep{spacebyte},
MambaByte~\citep{mambabyte}, Charformer~\citep{charformer}, and
CANINE~\citep{canine} explore hierarchical or tokenization-free alternatives,
using shortening, selective computation, state-space layers, learned soft
tokenization, or character-level encoders.
We do not claim superiority over those families; the contribution is a matched
twin comparison at one scale, together with the evaluation protocol needed to
make such a comparison fair.

\paragraph{Controlled tokenizer comparisons.}
A separate line of work holds the language model fixed and varies the
tokenizer.
\citet{ali2024tokenizer} train 24 models at 2.6B parameters with different
tokenizer algorithms and settings and find that the choice affects downstream
accuracy and training cost, and that fertility and parity are not always
predictive.
\citet{schmidt2024compression} test whether fewer tokens imply better models
across 64 models from 350M to 2.4B parameters and find that compression alone
does not explain downstream performance.
\citet{meister2026tokeval} proposes TokEval, a suite of intrinsic tokenizer
metrics, and validates it with controlled pretraining in which only the
tokenizer's data mixture, pretokenization and training algorithm change.
\citet{tao2024vocab} study how the compute-optimal vocabulary size grows with
compute.
All of these compare subword tokenizers with one another, and the training
budget is usually set in tokens or FLOPs.
We instead compare a tokenizer-free byte-patch model with one BPE twin and fix
the budget in raw bytes, so that both models see exactly the same text; the
question is whether the tokenizer can be removed, not which tokenizer to use.

\paragraph{Evaluating models with different tokenizers.}
Comparing models that segment text differently requires a common unit, and
BPB is the usual choice.
\citet{schaeffer2026myths} show that BPB is not tokenizer-agnostic (it factors
into tokens per byte times bits per token, and the factors do not cancel), that
lower BPB need not mean a better model, and that metrics proposed as
tokenizer-agnostic can disagree.
Our setting is exposed to the same issue, since our headline likelihood result
is in BPB.
We respond in three ways: both models score identical byte ranges with the
same left context (matched spans, Section~\ref{sec:protocol}); we report
downstream tasks as a second line of evidence that does not depend on BPB; and
we document concrete evaluation errors (Section~\ref{sec:pitfalls}) that each
changed a headline number in this byte-versus-BPE comparison.

\begin{table}[t]
\centering
\caption{Positioning against the closest prior studies (as described in each
paper's abstract). ``Held fixed'' is the main control in each design.}
\label{tab:related}
\small
\begin{tabular}{@{}>{\raggedright\arraybackslash}p{2.7cm}>{\raggedright\arraybackslash}p{4.2cm}>{\raggedright\arraybackslash}p{4.4cm}>{\raggedright\arraybackslash}p{2.5cm}@{}}
\toprule
Study & What is compared & Held fixed / controlled & Scale \\
\midrule
\citet{ali2024tokenizer} & subword tokenizer algorithms and settings & model size & 24 models, 2.6B \\
\citet{schmidt2024compression} & tokenization stages (pre-tokenization, vocabulary, segmentation) & model family & 64 models, 350M--2.4B \\
\citet{tao2024vocab} & vocabulary size & compute (IsoFLOPs) & 33M--3B \\
\citet{meister2026tokeval} & subword tokenizers (data mix, pretokenization, algorithm) & LM training setup & controlled sweep \\
\citet{blt} & learned byte patches vs.\ tokens & FLOPs, across a scaling sweep & up to 8B, 4T bytes \\
\textbf{This work} & fixed-patch byte model vs.\ BPE twin & raw bytes, byte order, optimizer recipe, bytes/step; scored byte spans & one pair, 350M \\
\bottomrule
\end{tabular}
\end{table}

\section{Models}
\label{sec:model}

Both models use width $D{=}1024$, $26$ layers and $16$ heads, so that the
global stack is identical in shape (Table~\ref{tab:configs}).
This section describes the two models module by module; shapes, initialization
and a forward sketch are collected in Appendix~\ref{app:arch}.

\subsection{\local: patch encoder, global Transformer, local decoder}

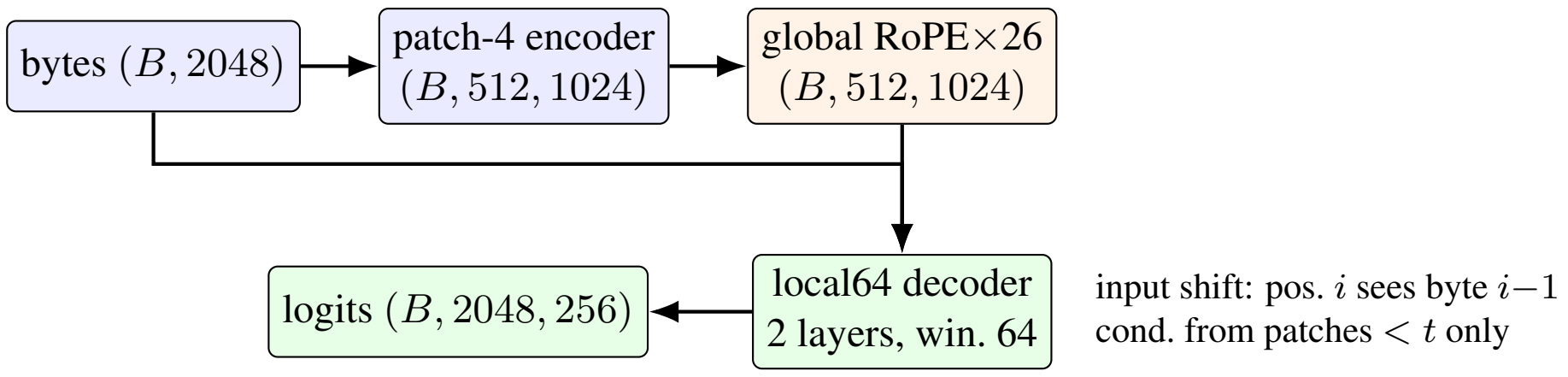
\begin{figure}[t]
\centering
\begin{tikzpicture}[
  font=\footnotesize,
  box/.style={draw, rounded corners=2pt, align=center, inner sep=3pt, minimum height=7mm},
  arr/.style={-Latex, thick}
]
\node[box, fill=blue!8] (bytes) {bytes $(B,2048)$};
\node[box, fill=blue!8, right=6mm of bytes] (enc) {patch-4 encoder\\$(B,512,1024)$};
\node[box, fill=orange!10, right=6mm of enc] (glob) {global RoPE$\times$26\\$(B,512,1024)$};
\node[box, fill=green!10, below=10mm of glob] (loc) {local64 decoder\\2 layers, win.\ 64};
\node[box, fill=green!10, left=8mm of loc] (out) {logits $(B,2048,256)$};
\draw[arr] (bytes) -- (enc);
\draw[arr] (enc) -- (glob);
\draw[arr] (glob) -- (loc);
\draw[arr] (loc) -- (out);
\draw[arr] (bytes.south) -- ++(0,-4mm) -| (loc.north);
\node[anchor=west, align=left, font=\scriptsize] at ($(loc.east)+(2mm,0)$)
  {input shift: pos.\ $i$ sees byte $i{-}1$\\
   cond.\ from patches $<t$ only};
\end{tikzpicture}
\caption{\local{} at 350M ($D{=}1024$). Bytes are grouped into non-overlapping
patches of length~4, encoded independently, mixed by a causal global
Transformer, then decoded by a two-layer sliding-window byte module.
Shapes are for a 2048-byte sequence.
\bpe{} is a standard decoder-only RoPE Transformer over an 8k vocabulary
(not shown).}
\label{fig:arch}
\end{figure}

Figure~\ref{fig:arch} summarizes \local.
A sequence of $T$ bytes is processed in four stages.

\paragraph{1. Patch encoder.}
Bytes are grouped into non-overlapping patches of length $p{=}4$.
Each byte is embedded (a $256\times1024$ table), a sinusoidal position within
the patch is added, and the four vectors pass through a LayerNorm and a
two-layer MLP ($D\to 2D\to D$ with GELU).
The four outputs are averaged into a single continuous vector per patch.
There is no attention in the encoder and no vector quantization,
so each patch embedding depends only on its own four bytes.

\paragraph{2. Global Transformer.}
The $T/4$ patch vectors go through an input LayerNorm and $26$ pre-LN
Transformer blocks with rotary position embeddings (RoPE), causal
self-attention, and a GELU MLP of hidden width $4D$, followed by a final
LayerNorm.
Attention runs over patches ($512$ positions for a 2048-byte sequence), so the
expensive stack sees a sequence four times shorter than the byte sequence.

\begin{figure}[t]
\centering
\begin{tikzpicture}[
  font=\footnotesize,
  pbox/.style={draw, rounded corners=2pt, fill=blue!8, align=center, minimum width=22mm, minimum height=9mm},
  state/.style={draw, rounded corners=2pt, fill=orange!10, minimum width=10mm, minimum height=6mm},
  arr/.style={-Latex, thick},
  carr/.style={-Latex, thick, dashed, orange!70!black}
]
\node[pbox] (p0) {patch 0\\bytes $0$--$3$};
\node[pbox, right=16mm of p0] (p1) {patch 1\\bytes $4$--$7$};
\node[pbox, right=16mm of p1] (p2) {patch 2\\bytes $8$--$11$};
\node[state, below=11mm of p0] (h0) {$h_0$};
\node[state, below=11mm of p1] (h1) {$h_1$};
\node[state, below=11mm of p2] (h2) {$h_2$};
\node[state, left=12mm of p0] (z) {$\mathbf{0}$};
\draw[arr] (p0) -- node[right, font=\scriptsize]{global} (h0);
\draw[arr] (p1) -- node[right, font=\scriptsize]{global} (h1);
\draw[arr] (p2) -- node[right, font=\scriptsize]{global} (h2);
\draw[carr] (z) -- (p0);
\draw[carr] (h0.north east) -- (p1.south west);
\draw[carr] (h1.north east) -- (p2.south west);
\end{tikzpicture}
\caption{Conditioning delay. The global state $h_t$ is computed causally from
patches $0,\dots,t$ and conditions the bytes of patch $t{+}1$ (dashed arrows);
patch~0 receives a zero vector. Inside a patch the decoder additionally sees the
preceding raw bytes through the one-byte input shift and its 64-byte window.}
\label{fig:cond}
\end{figure}
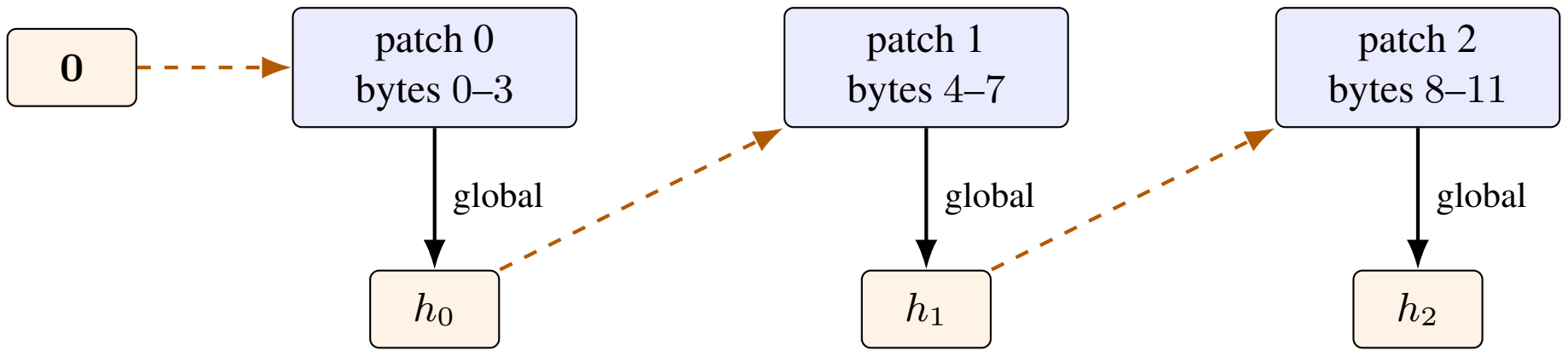

\paragraph{3. Conditioning with a one-patch delay.}
The global state is shifted by one patch before it is used.
The state $h_{t-1}$, which summarizes all bytes in patches before $t$, is
repeated across the four bytes of patch $t$; for the first patch the
conditioning vector is zero.
Thus the target bytes of patch $t$ never enter the computation that
conditions their own prediction (Figure~\ref{fig:cond}).

\paragraph{4. Local sliding-window decoder.}
The decoder input at byte position $i$ is the embedding of byte $i{-}1$ (a
learned start vector at $i{=}0$), concatenated with the conditioning vector for
that byte and projected back to width $D$ by a linear layer ($2D\to D$);
a sinusoidal position encoding is added.
Two pre-LN blocks follow, each with causal self-attention restricted to the last
64 bytes (16 heads) and a GELU MLP of hidden width $2D$.
A final LayerNorm and a linear head ($D\to256$) produce the byte logits.
The window is a sliding band over the byte sequence, not a block aligned to
patches, so a byte near the start of a patch attends to bytes in the previous
patch.
Stacking two such layers lets information flow up to $2{\cdot}64{-}1$ bytes
back.
The one-byte input shift guarantees that no layer can see the byte it predicts
or any later byte.
The decoder has its own byte embedding and output head; they are not shared with
the patch encoder.

The training objective is the mean cross-entropy over all scored bytes, and the
reported metric is bits per byte (Eq.~\ref{eq:bpb}):
\begin{equation}
\bpb
= \frac{1}{N\ln 2}\sum_{i=1}^{N}-\ln p(x_i\mid x_{<i}).
\label{eq:bpb}
\end{equation}

\subsection{\bpe: the matched token-level twin}

\bpe{} is a decoder-only Transformer with the same width, depth, head count,
RoPE blocks and MLP width as the \local{} global stack, but over $8000$ byte-level
BPE tokens instead of patches.
Its input embedding is tied to its output head; there is no dropout.
It is trained with next-token cross-entropy, ignoring the padding id.
For the byte-normalized likelihood we sum the target-token negative log-likelihoods
and divide by $\ln 2$ times the UTF-8 bytes of those scored target tokens
(Section~\ref{sec:protocol}).

\subsection{Parameters}

\begin{table}[t]
\centering
\caption{Parameter counts by module at 350M
(from the released parameter/FLOPs meta file).}
\label{tab:params}
\small
\begin{tabular}{@{}lrr@{}}
\toprule
Module & \local & \bpe \\
\midrule
Total & $351.3$M & $335.6$M \\
Non-embedding & $350.8$M & $327.4$M \\
Global / backbone (26$\times$ RoPE) & $327.4$M & $327.4$M \\
Local decoder / BPE embed (tied) & $19.4$M & $8.2$M \\
Patch encoder & $4.5$M & --- \\
\bottomrule
\end{tabular}
\end{table}

The shared 26-layer backbone accounts for $327.4$M parameters in both models.
\local{} adds a $19.4$M local decoder and a $4.5$M patch encoder; \bpe{} adds an
$8.2$M tied embedding table.
The byte model is therefore $4.7\%$ larger in total
(Table~\ref{tab:params}); we return to this in Section~\ref{sec:limits}.

\subsection{Differences from MEGABYTE}

Verified in code (Appendix~\ref{app:arch}): (i)~fixed patch size 4 with no VQ;
(ii)~local attention is a sliding 64-byte window that crosses patch boundaries,
not an intra-patch decoder; (iii)~one matched BPE twin rather than a patch-size
scaling study.
We do not claim inference-speed or scaling advantages beyond this experiment.

\section{Training setup}
\label{sec:setup}

\paragraph{Data.}
Both twins train on shuffled real\_v3 (85\% FineWeb-Edu~\citep{fineweb} /
10\% codeparrot-clean~\citep{codeparrot} /
5\% open-web-math~\citep{openwebmath}; train bytes $59.500$B / $7.000$B / $3.500$B;
train docs $12{,}300{,}716$ / $1{,}094{,}563$ / $494{,}592$).
Validation uses real\_v3 domain bins.
Documents are ordered by one fixed permutation (seed~0), shared by both runs, so
the two models see the same bytes in the same order.

\paragraph{Tokenizer.}
\bpe{} uses the matched 8k byte-level BPE trained on real\_v2 (vocab size
$8000$), fixed before these experiments and not tuned for either twin.

\paragraph{Matching.}
Both launch scripts use the same document permutation and the same byte bounds
on the packed real\_v3 train file.
BPE's batch $64\!\to\!37$ retune changes only the optimizer step size, not
document order.
Optimizer: AdamW ($\beta{=}(0.9,0.95)$, weight decay $0.1$, grad clip $1.0$),
bf16.
LR: cosine keyed to bytes (peak $3\!\times\!10^{-4}$, warmup $1\%$, floor
$10\%$).
\local{} runs at $65{,}536$ bytes/step (about $152.6$k steps for $10^{10}$
bytes); BPE packed at $\sim$114k~B/step then was retuned to $\sim$66k~B/step.
Keying the schedule to bytes rather than steps means that a change of batch size
does not change the learning-rate curve as a function of data seen.
One recipe difference remains: embedding initialization (Section~\ref{sec:limits}).

\paragraph{Why 512-token packs for BPE.}
The 350M \bpe{} twin packs concatenated window tokens into fixed length-$512$
sequences (mean $\approx 1760$~B per pack on the pretok stream).
That is the training context used throughout; evaluation past that pack is
discussed in Section~\ref{sec:pitfalls}.

\paragraph{Compute and checkpoints.}
Both runs used NVIDIA GB10 hardware, one twin per host
(wall-clock $\approx56.6$\,h for \local{} and $\approx60.6$\,h for \bpe{}
including pauses; Appendix~\ref{app:compute}).
Each twin is a single run to $10^{10}$ bytes (the \bpe{} run was resumed once
from a checkpoint at about $9.98\times10^{9}$ bytes after its token stream ran
out slightly early).

\paragraph{Checkpoints (frozen hashes).}
350M \local{} full-file \texttt{5080ed9b\ldots} /
model-tensor \texttt{33522128\ldots};
350M \bpe{} full-file \texttt{98f4a77c\ldots} /
model-tensor \texttt{53986bd8\ldots}
(see \path{research/FROZEN_RESULTS.md}).

\begin{table}[t]
\centering
\caption{Training configuration at 350M / $10^{10}$ bytes / real\_v3.
Train FLOPs/byte for \bpe{} uses seq${=}512$ (ratio BPE/\local{} $0.983$;
Appendix~\ref{app:compute}).}
\label{tab:configs}
\small
\begin{tabular}{@{}lcc@{}}
\toprule
 & \local & \bpe \\
\midrule
Params (total) & $351.3$M & $335.6$M \\
Params (non-emb.) & $350.8$M & $327.4$M \\
Layers / width / heads & $26/1024/16$ & $26/1024/16$ \\
Bytes trained & $10^{10}$ & $10^{10}$ \\
Train sequence & $2048$ B & $512$ tok \\
B/step (approx.) & $65{,}536$ & $114\mathrm{k}\!\to\!66\mathrm{k}$ \\
Train FLOPs/byte & $6.48\!\times\!10^{8}$ & $6.37\!\times\!10^{8}$ \\
Infer.\ FLOPs/byte & $2.16\!\times\!10^{8}$ & $2.12\!\times\!10^{8}$ \\
\bottomrule
\end{tabular}
\end{table}

\section{Evaluation protocol}
\label{sec:protocol}

\paragraph{Bits per byte.}
Both models are scored in bits per byte on the same validation documents.
For \local{} this is Eq.~\ref{eq:bpb} over the scored bytes.
For \bpe{} it is the sum of target-token NLLs divided by $\ln 2$ times the UTF-8
byte length of those scored target tokens, so that both models are measured per
byte of text.
The two quantities differ only in how the context is cut and which bytes are
scored, which is why the protocol below matters.

\paragraph{Primary: matched span.}
Document sliding with stride $512$\,B.
For \bpe{}, each window expands left until the next token would exceed
$\min(512$ tokens$,\,2048$\,B$)$.
For \local{}, the same left-context start byte is used, so both models score
identical byte ranges.
Mean effective context: text/code/math $1671$/$900$/$1538$\,B.
Paired bootstrap over documents ($10^{4}$ resamples); we report
$\Delta{=}\mathrm{BPB}_{\bpe}-\mathrm{BPB}_{\local}$ with $95\%$ CIs.
Documents are scored as they appear in the validation bins: the separator byte \texttt{0x1E} that delimits documents is dropped and nothing is prepended (the first-position audit that prepends it is in Appendix~\ref{app:alt}).

\paragraph{Why shared-cap and own-context are not primary.}
Shared-cap ($\min(512$ tok$,\,1760$\,B$)$ / local $1760$\,B) and own-context
($512$ tok / local $2048$\,B) leave unequal byte spans on code
($\sim$900\,B for \bpe{} vs $\sim$1630--1866\,B for \local{}).
Those protocols show a code tie or \local{} win that disappears when spans are
equalized (Appendix~\ref{app:alt}).
We therefore treat them as diagnostics, not as the headline comparison.

\paragraph{Downstream tasks.}
We evaluate ten tests on six benchmarks: HellaSwag~\citep{hellaswag},
PIQA~\citep{piqa} and ARC-Easy~\citep{arc}
(each with raw and length-normalized scoring), LAMBADA~\citep{lambada}, fact200, and CharBench
with 5-shot prompting and with equal fine-tuning.
For the multiple-choice sets, the raw score of an answer is the total
log-probability of that answer given the context, and the length-normalized score
divides it by the UTF-8 byte length of the answer string, so both models are
normalized by the same unit; the prediction is the argmax over answers.
LAMBADA is scored by exact match of the greedy continuation of the final word
(the target includes its leading space); fact200 by alias-normalized exact match
of a greedy answer after a few-shot prompt; and CharBench by string match of the
first generated line against the gold answer.
CharBench has twelve character-level tasks (first/last/nth character, count,
reverse, length, case, edit, sort, palindrome) with $50$ test items per task
($n{=}600$); in the equal fine-tuning setting both models are tuned on $200$
training items per task with the same learning rate, which was selected on a
separate validation split.
Datasets, splits and item counts are listed in Appendix~\ref{app:downstream}.

\paragraph{Statistics.}
Accuracies carry Wilson $95\%$ intervals.
Model pairs are compared with an exact McNemar test on paired item outcomes,
with Holm correction across all ten tests; we report significance at
$\alpha{=}0.05$ and, as a stricter check, at $0.01$.

\section{Results}
\label{sec:results}

\begin{table}[t]
\centering
\caption{350M @ 10B matched-span BPB (primary).
Positive $\Delta$ means \local{} is better.
Relative gap $=(\mathrm{BPB}_{\bpe}-\mathrm{BPB}_{\local})/\mathrm{BPB}_{\bpe}$.}
\label{tab:bpb}
\small
\begin{tabular}{@{}lcccc@{}}
\toprule
Domain & \local & \bpe & $\Delta$ $[95\%]$ & Rel.\ \% \\
\midrule
overall & $0.9634$ & $0.9155$ & $-0.0479$ $[-0.0506,-0.0453]$ & $-5.2$ \\
text & $0.9907$ & $0.9448$ & $-0.0459$ $[-0.0486,-0.0434]$ & $-4.9$ \\
code & $0.6563$ & $0.6003$ & $-0.0560$ $[-0.0683,-0.0439]$ & $-9.3$ \\
math & $1.1023$ & $1.0388$ & $-0.0635$ $[-0.0749,-0.0530]$ & $-6.1$ \\
\bottomrule
\end{tabular}
\end{table}

\begin{figure}[t]
\centering
\includegraphics[width=0.98\linewidth]{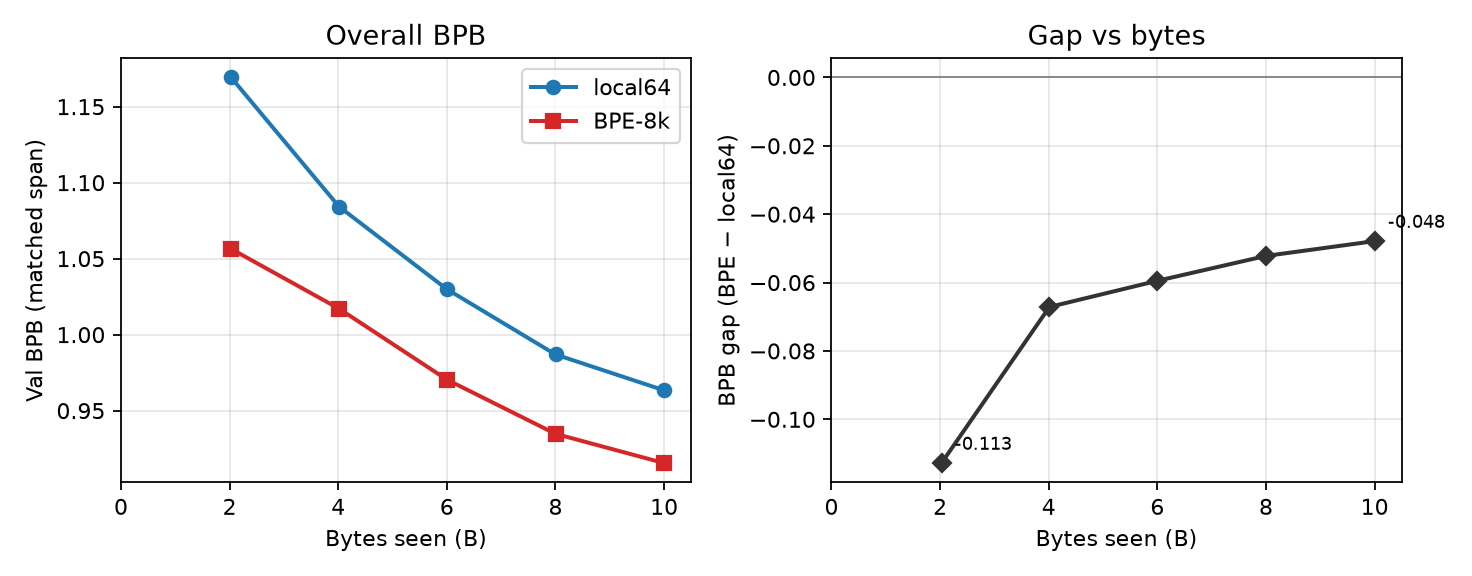}
\caption{Left: overall matched-span BPB vs bytes seen.
Right: gap $\mathrm{BPB}_{\bpe}-\mathrm{BPB}_{\local}$, from $-0.113$ at 2B to
$-0.048$ at 10B.
The 10B gap equals Table~\ref{tab:bpb} overall $\Delta$.
Intermediate checkpoints are not annealed.}
\label{fig:curve}
\end{figure}

\paragraph{Likelihood.}
\bpe{} is significantly better on all four domains under matched spans
(Table~\ref{tab:bpb}; relative gaps $-5.2\%$, $-4.9\%$, $-9.3\%$, $-6.1\%$
(overall, text, code, math)).
None of the four confidence intervals includes zero.
Figure~\ref{fig:curve} shows the gap shrinks with bytes but stays negative.
At the checkpoints we evaluated, the overall gap is $-0.113$ (2B),
$-0.067$ (4B), $-0.059$ (6B), $-0.052$ (8B) and $-0.048$ (10B); the decrease
slows over the last three checkpoints, and we do not extrapolate beyond $10^{10}$
bytes.

\begin{table}[t]
\centering
\caption{Downstream accuracy (350M @ 10B) with Wilson intervals.
Bold: Holm-significant at $\alpha{=}0.05$ across all ten McNemar tests.}
\label{tab:down}
\small
\begin{tabular}{@{}lcc@{}}
\toprule
Task & \local & \bpe \\
\midrule
HellaSwag raw & $0.2846$ $[0.2759$--$0.2935]$ & $0.2846$ $[0.2759$--$0.2935]$ \\
HellaSwag len-norm & $0.3030$ $[0.2941$--$0.3121]$ & $0.3005$ $[0.2916$--$0.3096]$ \\
PIQA raw & $0.5887$ $[0.5660$--$0.6110]$ & $\mathbf{0.6143}$ $[0.5918$--$0.6362]$ \\
PIQA len-norm & $0.5756$ $[0.5529$--$0.5980]$ & $\mathbf{0.6034}$ $[0.5808$--$0.6255]$ \\
ARC-Easy raw & $0.4456$ $[0.4053$--$0.4866]$ & $0.4895$ $[0.4486$--$0.5304]$ \\
ARC-Easy len-norm & $0.3281$ $[0.2908$--$0.3677]$ & $\mathbf{0.3842}$ $[0.3452$--$0.4248]$ \\
LAMBADA & $0.2360$ $[0.2246$--$0.2478]$ & $0.2401$ $[0.2286$--$0.2519]$ \\
fact200 & $0.0700$ $[0.0422$--$0.1141]$ & $\mathbf{0.1650}$ $[0.1200$--$0.2227]$ \\
CharBench few-shot & $0.2083$ $[0.1777$--$0.2426]$ & $0.1783$ $[0.1498$--$0.2110]$ \\
CharBench equal-FT & $\mathbf{0.5367}$ $[0.4967$--$0.5762]$ & $0.4000$ $[0.3616$--$0.4397]$ \\
\bottomrule
\end{tabular}
\end{table}

\begin{table}[t]
\centering
\caption{McNemar exact $p$ and Holm-adjusted $p$ across ten tests.
Bold winners in Table~\ref{tab:down} have $p_{\mathrm{Holm}}{<}0.05$.}
\label{tab:holm}
\small
\begin{tabular}{@{}lccccc@{}}
\toprule
Test & $p_{\mathrm{raw}}$ & $p_{\mathrm{Holm}}$ & Winner & $0.05$ & $0.01$ \\
\midrule
HellaSwag raw & $1.000$ & $1.000$ & --- & no & no \\
HellaSwag len-norm & $0.483$ & $1.000$ & --- & no & no \\
PIQA raw & $0.0056$ & $0.034$ & \bpe{} & yes & no \\
PIQA len-norm & $0.0020$ & $0.014$ & \bpe{} & yes & no \\
ARC-Easy raw & $0.068$ & $0.339$ & --- & no & no \\
ARC-Easy len-norm & $0.0016$ & $0.013$ & \bpe{} & yes & no \\
LAMBADA & $0.479$ & $1.000$ & --- & no & no \\
fact200 & $0.00088$ & $0.0079$ & \bpe{} & yes & yes \\
CharBench few-shot & $0.127$ & $0.506$ & --- & no & no \\
CharBench equal-FT & $<10^{-6}$ & $<10^{-5}$ & \local{} & yes & yes \\
\bottomrule
\end{tabular}
\end{table}

\paragraph{Downstream.}
At $\alpha{=}0.05$ after Holm correction, \bpe{} wins PIQA raw, PIQA
length-normalized, ARC-Easy length-normalized, and fact200; \local{} wins
CharBench equal fine-tuning.
Only fact200 and that fine-tune remain at $0.01$.
The two HellaSwag rows and LAMBADA are ties.
HellaSwag accuracy is close to chance at this size (the two models agree on
$86.6\%$ of items, so identical raw accuracy is a coincidence of totals, not a
duplicated checkpoint), and we do not treat it as informative.

\section{Analysis}
\label{sec:analysis}

This section interprets the results.
Statements marked \emph{Hypothesis} are explanations we find plausible but did
not test with ablations.

\paragraph{Code under matched context.}
Under matched spans, \bpe{} leads on code ($\Delta{=}{-}0.056$, rel.\ $-9.3\%$).
Under own-context, \local{} sees $\sim$1866\,B of code context versus $\sim$900\,B
for \bpe{}, and the code gap flips to a \local{} win ($\Delta{=}{+}0.024$).
Shared-cap leaves a code tie with the same span mismatch
(Appendix~\ref{app:alt}).
\emph{Hypothesis:} \local{}'s advantage under own-context is a context-length
effect, not a domain effect favoring bytes on code.
The practical consequence is that the same two checkpoints give opposite code
rankings under two plausible evaluation protocols, which is why we make the
span-matched protocol primary.

\paragraph{Hypothesis A (character tasks favor bytes).}
CharBench equal fine-tuning favors \local{} after Holm correction
($53.7\%$ vs $40.0\%$).
\emph{Hypothesis:} counting, indexing, reversing, and case transforms align more
directly with byte state than with BPE merges.
Notably, the few-shot version of the same benchmark is a tie, so the advantage
appears only after fine-tuning.

\paragraph{Hypothesis B (subword items favor BPE).}
fact200 favors \bpe{} ($16.5\%$ vs $7.0\%$ for \local), as do PIQA and ARC-Easy
length-normalized at Holm $0.05$.
\emph{Hypothesis:} named entities and short answer choices are stored more
sample-efficiently as whole subwords under the same byte budget.

\paragraph{Cost.}
The byte model is not cheaper in this comparison.
It has $4.7\%$ more parameters and, under our matmul count, about $2\%$ more
training FLOPs per byte (ratio $0.983$; Table~\ref{tab:configs}), and the BPE twin still has
the lower bits per byte.
The comparison is data- and optimizer-matched, not FLOPs-matched, and we make
no inference-speed claim.

\paragraph{What the results do not show.}
They do not show that byte-patch models are worse than BPE models in general:
the result is for one size, one seed, one 8k vocabulary, and one BPE training
context length (Section~\ref{sec:limits}).
In particular, a BPE model trained on longer sequences, a larger vocabulary, or
a different patch size for the byte model, or a \local{} trained with the same embedding initialization as \bpe{}, could change the ranking, and we did
not test those.

\section{Evaluation pitfalls}
\label{sec:pitfalls}

During the study we found several evaluation errors, each of which changed a
number we would otherwise have reported.
We list them as a checklist for comparisons of byte-level and token-level
models.

\begin{itemize}[leftmargin=*,itemsep=2pt]
\item \textbf{Bits/token as BPB.}
An internal BPE path divided mean token cross-entropy by $\ln 2$ only
($3.1979$ overall). Dividing by a byte count is required for BPB.
\item \textbf{Wrong BPE window denominator.}
An internal path divided target-token NLLs by the UTF-8 byte count of the
re-encoded decoded window (a shorter string than the scored target-token
bytes, so the denominator was too small and BPB too high).
Wrong overall $0.9792$; corrected $0.9774$ (uncapped windows).
\item \textbf{First-position asymmetry.}
Without a leading separator, \local{} can score byte~0 from a learned start
state while BPE leaves the first token unscored.
Prepending \texttt{0x1E} recovers those bytes; paired $\Delta$ moves by at most
$0.001$ (Appendix~\ref{app:alt}).
\item \textbf{LAMBADA first character.}
\texttt{choices[0]} on a string with a trailing space on the context:
\local{} $0.2926\!\to\!0.2360$; \bpe{} $0.0930\!\to\!0.2401$.
\item \textbf{\texttt{torch.compile} keys.}
Loaders strip \texttt{\_orig\_mod.} prefixes ($305$ on \local{}, $0$
missing/unexpected).
\item \textbf{Bytes-per-step mismatch.}
BPE was retuned from $\sim$114k to $\sim$66k~B/step to match \local{} at
$65{,}536$.
\item \textbf{Context-length extrapolation.}
\bpe{} trains at $512$ tokens.
Under matched spans, BPE code BPB is $0.600$; the same checkpoint on fixed
2048-byte windows reports $0.843$ (labelled not comparable;
Appendix~\ref{app:alt}).
Because the 8k vocabulary compresses code at about $1.95$ bytes per token, $512$
tokens cover only about $900$\,B of code, so a ``2048-byte'' window is far
outside the trained range.
\item \textbf{Unequal-context comparisons.}
Shared-cap / own-context give \local{} more code bytes than \bpe{}; matched
spans remove that confound and flip the code ranking.
\end{itemize}

\section{Limitations}
\label{sec:limits}

We report one size and one seed at 350M.
\bpe{} was trained at $512$ tokens, so matched-span code context is
$\sim$900\,B; a longer-context BPE was not tested.
\local{} was trained at $2048$-byte contexts and is scored at $\sim$900\,B on
code under matched spans (within its trained range, but possibly
disadvantaged relative to training).
Neither matched-span nor own-context is perfectly neutral on code: the former
equalizes spans at BPE's shorter code context, while the latter gives \local{}
more code bytes.
HellaSwag is near chance at this size and is uninformative for the comparison.
Our main likelihood metric is BPB, which is not tokenizer-agnostic and need
not track downstream quality~\citep{schaeffer2026myths}; matched spans
equalize the scored bytes and context but do not remove this concern, which is
why we report downstream tasks alongside it and limit our claims to this setup.
One 8k vocabulary was chosen before the experiments and not tuned; larger
vocabularies were not tested.
Initialization is not identical across the twins: \bpe{} embeddings are drawn from $\mathcal{N}(0,0.02^2)$, while the \local{} byte embeddings (patch encoder and decoder) keep the framework default $\mathcal{N}(0,1)$; residual-output scaling is applied to both. We did not test the effect, so part of the measured gap could come from this recipe difference rather than from the architecture.
Parameter counts differ within the setting (Table~\ref{tab:params}), and the
FLOPs comparison is a matmul-only estimate (Appendix~\ref{app:compute}).
We did not ablate the local decoder (window size, depth) or the patch size.
We make no inference-speed or scaling claims beyond 350M / $10^{10}$ bytes.
Preliminary 120M twins used a different BPE packing pipeline and are excluded
from the main conclusions (Appendix~\ref{app:120m}).

\section{Conclusion}
\label{sec:conclusion}

Under matched data and matched scored spans at 350M ($10^{10}$ bytes),
\bpe{} leads overall/text/code/math (relative gaps $-5.2\%$, $-4.9\%$,
$-9.3\%$, $-6.1\%$ (overall, text, code, math)).
Character-task fine-tuning favors the byte model; factoid and some
multiple-choice tasks favor BPE after Holm correction.
This paper does not show that byte-patch models are better than BPE in
general---only that, at this size and protocol, the matched BPE twin wins
likelihood when context is equalized.
We release the checkpoints, hash-locked evaluation suites and the evaluation
checklist so that these comparisons can be repeated and extended to longer
BPE contexts, more seeds, and other patch sizes.

\appendix
\section{Hyperparameters}
\label{app:hparams}
AdamW $\beta=(0.9,0.95)$, weight decay $0.1$, grad clip $1.0$, bf16, dropout $0$.
LR: cosine on bytes, peak $3\!\times\!10^{-4}$, warmup $1\%$, min fraction $0.1$.
\local{}: sequence $2048$ B, $65{,}536$ B/step, global stack on $512$ patches.
CharBench equal-FT LR $5\!\times\!10^{-5}$ for both models.
Scaled residual init: \bpe{} embedding std $0.02$; residual attention-out and
MLP-out std $0.02/\sqrt{2L}$ with $L{=}26$ ($\approx0.00277$) in both global
stacks; the \local{} decoder's two layers are scaled separately
($0.02/\sqrt{4}{=}0.01$).
The \bpe{} tied input/output embedding is explicitly set to $\mathcal{N}(0,0.02^{2})$
in \path{scripts/train_bpe_matched350m.py}; \local{} patch-encoder and decoder
byte embeddings keep PyTorch's default $\mathcal{N}(0,1)$ (only residual
attn/MLP outs are re-scaled in \path{scripts/train_local64_ddp.py}).
Initialization of the remaining parameters follows the released training
scripts (\path{research/ARCHITECTURE_SPEC.md}, Section~4).

\section{Data manifest}
\label{app:data}
\textbf{real\_v3:} 85/10/5; train bytes
$59{,}500{,}002{,}824$ / $7{,}000{,}003{,}274$ / $3{,}500{,}012{,}596$;
train docs $12{,}300{,}716$ / $1{,}094{,}563$ / $494{,}592$.

\section{CharBench tasks}
\label{app:charbench}
Twelve tasks (first/last/nth char, count, reverse, length, case, edit, sort,
palindrome).
Defaults: 5-shot, 50 test per task ($n{=}600$), 200 train per task for equal FT.
Suite lock sha256:\\
{\ttfamily\footnotesize becda801323a53aa4e558e26b50544356532f487e3717d693a450fe0209a02a8}.

\section{fact200}
\label{app:fact200}
$n{=}200$; lock sha256:\\
{\ttfamily\footnotesize 545a7dcde7c65b03fb22790359eb20196732a64a55a39480d3e4de7d600a1a16}.

\section{Alternative protocols}
\label{app:alt}

\begin{table}[h]
\centering
\caption{Alternate 350M context protocols (not primary).
Matched-span numbers are in Table~\ref{tab:bpb}.}
\label{tab:alt350}
\small
\begin{tabular}{@{}lcc@{}}
\toprule
Domain & shared cap $\Delta$ $[95\%]$ & own ctx $\Delta$ $[95\%]$ \\
\midrule
overall & $-0.0400$ $[-0.0437,-0.0359]$ & $-0.0378$ $[-0.0417,-0.0336]$ \\
text & $-0.0456$ $[-0.0484,-0.0431]$ & $-0.0445$ $[-0.0472,-0.0421]$ \\
code & $+0.0147$ $[-0.0034,+0.0308]$ & $+0.0243$ $[+0.0056,+0.0409]$ \\
math & $-0.0531$ $[-0.0626,-0.0461]$ & $-0.0481$ $[-0.0595,-0.0404]$ \\
\bottomrule
\end{tabular}
\end{table}

\begin{table}[h]
\centering
\caption{Mean effective context bytes (real\_v3 val).
Matched span is identical for both models.}
\label{tab:effctx}
\small
\begin{tabular}{@{}lccccc@{}}
\toprule
Domain & shared \bpe & shared \local & own \bpe & own \local & matched \\
\midrule
text & $1525$ & $1548$ & $1671$ & $1756$ & $1671$ \\
code & $900$ & $1630$ & $900$ & $1866$ & $900$ \\
math & $1468$ & $1618$ & $1538$ & $1856$ & $1538$ \\
\bottomrule
\end{tabular}
\end{table}

\begin{table}[h]
\centering
\caption{Uncapped fixed 2048-byte windows at 350M @ 10B.
\textbf{Not comparable:} BPE beyond trained length on code/math.}
\label{tab:uncapped350}
\small
\begin{tabular}{@{}lccc@{}}
\toprule
Domain & \local & \bpe & $\Delta$ $[95\%]$ \\
\midrule
overall & $1.0011$ & $0.9774$ & $-0.0237$ $[-0.0271,-0.0203]$ \\
text & $1.0291$ & $0.9845$ & $-0.0446$ $[-0.0461,-0.0432]$ \\
code & $0.6692$ & $0.8433$ & $+0.1741$ $[+0.1566,+0.1925]$ \\
math & $1.1745$ & $1.1111$ & $-0.0634$ $[-0.0699,-0.0572]$ \\
\bottomrule
\end{tabular}
\end{table}

First-position audit with \texttt{0x1E} prepend (shared-cap setting): paired
$\Delta$ overall/text/code/math
$-0.0398/-0.0454/+0.0157/-0.0537$
(vs Table~\ref{tab:alt350} $-0.0400/-0.0456/+0.0147/-0.0531$).

\section{Context lengths in tokens}
\label{app:tok}
\begin{table}[h]
\centering
\caption{BPE-8k tokens per full 2048-byte window on real\_v3 val
(trailing partial windows excluded).}
\label{tab:tokperwin}
\small
\begin{tabular}{@{}lcccc@{}}
\toprule
Domain & mean & p50 & max & $>512$? \\
\midrule
text & $531$ & $518$ & $966$ & yes \\
code & $1154$ & $1163$ & $1604$ & yes \\
math & $645$ & $617$ & $1100$ & yes \\
\bottomrule
\end{tabular}
\end{table}

\section{Compute}
\label{app:compute}
Hardware: NVIDIA GB10 --- \local{} twin on one host, \bpe{} on the other.
\local{} 350M wall $\approx 56.6$\,h.
\bpe{} 350M wall $\approx 60.6$\,h including pauses.

\paragraph{Counting rules.}
One multiply and one add each count as one FLOP; layer norms, softmax and GELU
are excluded; training is counted as $3\times$ the forward pass.
Counts cover matrix multiplications only, so they are estimates and not
measured hardware FLOPs.

\paragraph{\local{} worked example.}
With $d{=}1024$ and a $2048$-byte sequence, the global stack runs on
$2048/4{=}512$ patches.
One global layer costs
$8d^2+4d\cdot512+16d^2=27{,}262{,}976$ FLOPs per patch position.
Over $26$ layers, amortized over four bytes per patch, this is
$26\cdot27{,}262{,}976/4=177{,}209{,}344$ forward FLOPs per byte.
One local-decoder layer (window $64$, MLP factor $2$) costs
$8d^2+4d\cdot64+4\cdot2\cdot d^2=17{,}039{,}360$ FLOPs per byte; two layers
plus the fuse projection ($4d^2{=}4{,}194{,}304$) and the output head
($2d\cdot256{=}524{,}288$) give $38{,}797{,}312$ forward FLOPs per byte.
The training cost is
\[
3\cdot(177{,}209{,}344+38{,}797{,}312)=648{,}019{,}968\approx6.48\times10^{8}
\]
FLOPs per byte, and inference is $216{,}006{,}656\approx2.16\times10^{8}$.

\paragraph{\bpe{}.}
Under the same rules, \bpe{} at train pack $\mathrm{seq}{=}512$ includes the
$26$ transformer layers plus the tied $8000$-way output projection
($2d\cdot8000$ FLOPs per token).
Converting with the frozen val mean of $3.415$ bytes/token gives
$6.37\!\times\!10^{8}$ training FLOPs per byte (ratio BPE/\local{} $0.983$)
and $2.12\!\times\!10^{8}$ inference FLOPs per byte.
The \local{} patch-encoder MLP is \emph{not} included in either Table~\ref{tab:configs}
count (it has no BPE analogue; including it would raise \local{} alone to
$6.73\!\times\!10^{8}$ training FLOPs/byte).
The released meta file stores a different estimate that assumes a $2048$-token
sequence ($7.66\times10^{8}$ FLOPs per byte); it is not the training
configuration and is not used in the paper's comparisons.

\section{FLOPs counting and architecture notes}
\label{app:arch}
Full module list, shapes, init, and MEGABYTE differences:
\path{research/ARCHITECTURE_SPEC.md}.

\begin{table}[h]
\centering
\caption{Parameters by module at 350M (counts from the released meta file).}
\label{tab:modparams}
\small
\begin{tabular}{@{}llr@{}}
\toprule
Model & Module & Parameters \\
\midrule
\local & 26 RoPE blocks (global) & $327{,}395{,}328$ \\
\local & local decoder (total) & $19{,}425{,}536$ \\
\local & \quad 2 sliding-window blocks & $16{,}799{,}744$ \\
\local & \quad fuse $(2D\to D)$ & $2{,}098{,}176$ \\
\local & \quad head $(D\to256)$ & $262{,}400$ \\
\local & \quad byte embedding & $262{,}144$ \\
\local & \quad LayerNorm and start vector & $3{,}072$ \\
\local & patch encoder (total) & $4{,}461{,}568$ \\
\local & input / final LayerNorm & $2{,}048+2{,}048$ \\
\local & \textbf{total} & $\mathbf{351{,}286{,}528}$ \\
\midrule
\bpe & 26 RoPE blocks & $327{,}395{,}328$ \\
\bpe & embedding (tied head), $8000\times1024$ & $8{,}192{,}000$ \\
\bpe & final LayerNorm & $2{,}048$ \\
\bpe & \textbf{total} & $\mathbf{335{,}589{,}376}$ \\
\bottomrule
\end{tabular}
\end{table}

\begin{lstlisting}[caption={Forward sketch (\local{})},label={lst:fwd}]
# bytes_x: (B, T), T multiple of 4 preferred
emb = encode_patches(bytes_x)          # (B, T/4, D)
h   = patch_transformer(emb)           # causal RoPE stack
cond = shift_patches(h)                # patch t <- state < t
logits = local64_decode(cond, bytes_x) # shift-1 + win 64
loss = CE(logits, bytes_x)
\end{lstlisting}

\begin{lstlisting}[caption={BPB for both models},label={lst:bpb}]
# local: sum -log p(byte_i) / (ln2 * N_bytes)
# BPE:   sum token_NLL_nats / (ln2 * utf8_bytes(scored tokens))
BPB = total_nats / (ln(2) * scored_bytes)
\end{lstlisting}

\section{Preliminary 120M results}
\label{app:120m}

Old 120M twins (real\_v2, $2.5\!\times\!10^{9}$ bytes, two seeds) used a
different BPE pipeline: per-window tokenization of 2048-byte chunks
($\sim$650 tokens), no BPE scaled residual init, and step-keyed LR.
The old BPE twin is defective relative to the 350M packing and init recipe
(\path{research/diagnosis_120m.md}), so the numbers below are
\textbf{not} used for any main-text conclusion.

\begin{table}[h]
\centering
\caption{120M matched-span BPB on real\_v3 val (seed~0): local64, old BPE, and
packed-512 BPE control.
Source: \texttt{control\_120m\_bpe\_packed512\_eval.json}.}
\label{tab:120mctrl}
\small
\begin{tabular}{@{}lccc@{}}
\toprule
Domain & local64 s0 & old BPE s0 & packed-512 control \\
\midrule
overall & $1.2131$ & $1.2254$ & $1.0721$ \\
text & $1.2580$ & $1.2672$ & $1.1153$ \\
code & $0.7732$ & $0.8095$ & $0.6531$ \\
math & $1.3243$ & $1.3403$ & $1.1710$ \\
\bottomrule
\end{tabular}
\end{table}

Locked real\_v2 gate (training val; different protocol): local64 $1.2167$,
BPE $1.2340$.
Under 2048-B document slide the packed control's code BPB collapses
out-of-pack (see the same control eval artifact).

\section{Reproducibility}
\label{app:repro}
Release (code, evaluation suites, checkpoints and validation data):
{\raggedright
\begin{itemize}[leftmargin=*,itemsep=2pt]
\item GitHub:\\
  \url{https://github.com/AsadullahMaliyawala/bytelm-local64}
\item local64 weights:\\
  \url{https://huggingface.co/Asadullah1569/bytelm-local64-350m}
\item BPE weights:\\
  \url{https://huggingface.co/Asadullah1569/bytelm-bpe8k-350m-baseline}
\item Validation data:\\
  \url{https://huggingface.co/datasets/Asadullah1569/bytelm-local64-val}
\item Licences: code and weights Apache-2.0; training and validation data keep
  their upstream licences (FineWeb-Edu and OpenWebMath: ODC-By~1.0 plus Common
  Crawl terms of use; codeparrot-clean: mixed per-file licences).
\end{itemize}
}
All \path{research/} and \path{scripts/} paths cited in this paper are relative
to the root of the GitHub repository above.
Locked numbers: \path{research/FROZEN_RESULTS.md},
\path{research/paper_v10_numbers.json}.
Matched-span BPB: \path{research/slide_350m_matched_span.json}.
Curve: \path{research/curve_350m_matched_span.json}.
Downstream / Holm: \path{research/downstream_items.json},
\path{research/downstream_holm.json}.
Architecture: \path{research/ARCHITECTURE_SPEC.md}.
Eval-code hashes at freeze: see \path{research/FROZEN_RESULTS.md}.

\section{Downstream task details}
\label{app:downstream}

\begin{itemize}[leftmargin=*,itemsep=3pt]
\item \textbf{HellaSwag:} \path{Rowan/hellaswag}, validation, $n{=}10{,}042$.
  Argmax of the answer log-probability (raw, and per UTF-8 byte of the ending).
\item \textbf{PIQA:} \path{ybisk/piqa}, validation, $n{=}1{,}838$.
  Argmax of the solution log-probability (raw, and per UTF-8 byte).
\item \textbf{ARC-Easy:} \path{allenai/ai2_arc} (ARC-Easy), validation,
  $n{=}570$. Argmax of the choice log-probability (raw, and per UTF-8 byte).
\item \textbf{LAMBADA:} \path{EleutherAI/lambada_openai}, test, $n{=}5{,}153$.
  Exact match of the greedy continuation of the final word.
\item \textbf{fact200:} local suite, hash-locked (Appendix~\ref{app:fact200}),
  $n{=}200$. Alias-normalized exact match after a few-shot prompt.
\item \textbf{CharBench:} local suite, hash-locked (Appendix~\ref{app:charbench}),
  $n{=}600$ ($12$ tasks $\times$ $50$). First-line string match against the gold
  answer; the validation split is used only to select the equal fine-tuning
  learning rate.
\end{itemize}

ARC-Easy uses the $570$-item validation split, so its intervals are wide
(Table~\ref{tab:down}).
HellaSwag raw accuracy is identical for the two models ($0.2846$); the models
agree on $86.6\%$ of items ($8{,}700$ of $10{,}042$), so the identical total is a
coincidence and not evidence of a shared checkpoint
(\path{research/matched_eval_audit_summary.json}).

\section*{Data and code availability}
Code, checkpoints, evaluation suites and validation data are listed in
Appendix~\ref{app:repro}.
Code and weights are Apache-2.0; training and validation data keep their
upstream licences (FineWeb-Edu and OpenWebMath: ODC-By~1.0 plus Common Crawl
terms of use; codeparrot-clean: mixed per-file licences).

\section*{Use of AI tools}
The authors used AI assistants (Claude and Cursor) to write and run experiment
code, orchestrate evaluations, and draft and edit this paper.
The authors designed the study, reviewed all results, and are responsible for all
claims.

\section*{Author contributions}
A.M. designed the study, ran the experiments and wrote the paper.
A.L. performed calculations and analysis.

\bibliographystyle{plainnat}
\bibliography{refs}

\end{document}